# Developing and Using Special-Purpose Lexicons for Cohort Selection from Clinical Notes

Samarth Rawal, Ashok Prakash, Soumya Adhya, Sidharth Kulkarni,
Saadat Anwar, Chitta Baral, Murthy Devarakonda[1]

Arizona State University
Tempe, AZ

## Abstract

Background and Significance: Selecting cohorts for a clinical trial typically requires costly and time-consuming manual chart reviews resulting in poor participation. To help automate the process, National NLP Clinical Challenges (N2C2) conducted a shared challenge by defining 13 criteria for clinical trial cohort selection and by providing training and test datasets. This research was motivated by the N2C2 challenge.

Methods: We broke down the task into 13 independent subtasks corresponding to each criterion and implemented subtasks using rules or a supervised machine learning model. Each task critically depended on knowledge resources in the form of task-specific lexicons, for which we developed a novel model-driven approach. The approach allowed us to first expand the lexicon from a seed set and then remove noise from the list, thus improving the accuracy.

Results: Our system achieved an overall F measure of 0.9003 at the challenge, and was statistically tied for the first place out of 45 participants. The model-driven lexicon development and further debugging the rules/code on the training set improved overall F measure to 0.9140, overtaking the best numerical result at the challenge.

Discussion: Cohort selection, like phenotype extraction and classification, is amenable to rule-based or simple machine learning methods, however, the lexicons involved – such as medication names or medical terms referring to a medical problem – critically determine the overall accuracy. Automated lexicon development has the potential for scalability and accuracy.

## Background and Significance

Participation in clinical trials is alarmingly low, even though they are a vital part of medical research for developing new methods, drugs, and instruments for prevention, diagnosis, screening, treatment, and quality of life improvement. A 2010 study by the US Institute of Medicine [1] observed that only about 3% of adult cancer patients participate in clinical trials, and 40% of trials failed to achieve minimum patient enrollment. This is, in part, because clinical trials often have complex criteria that cannot be simply assessed using a database query on the structured (coded) part of a patient's record. It requires a careful review of the clinical narratives in a patient's record. It is a time-consuming process requiring advanced medical training, and so often researchers are limited to patients who either seek out the trials or who have been referred by their physicians.

---

[1] Contact author, email: **murthy.devarakonda@asu.edu**

Using researchers' familiarity or referrals alone can result in selection bias towards certain populations (e.g. people who can afford regular care or people who exclusively use free clinics), which in turn can bias the results of the study [2][3]. Indeed, there are many studies that point out lack of trials participation from minority groups and rural areas. An automated analysis of clinical notes, which can be eventually included in EHR systems, can reduce the time and cost it takes to identify patients, and can potentially help remove the biases in participant selection. Due to the potential complexity of the selection criteria there is a need for advanced natural language processing methods for such an automated selection.

National NLP Clinical Challenges (N2C2), which has evolved from the well-known i2b2 (Informatics for Integrating Biology and the Bedside) NLP Shared Task challenges, has organized a shared challenge to address the question "Can NLP systems use narrative medical records to identify which patients meet selection criteria for clinical trials?" [4]. The task required NLP systems to determine if each patient meets a specific list of 13 selection criteria, using 3-5 clinical notes provided for a patient, and determine if the patient meets, does not meet, each criterion. The organizers have developed the eligibility criteria based on real clinical trials. The criteria included patients' medications, past medical histories, and whether certain events have occurred in a specified timeframe in the patients' records.

The selection criteria for matching patients to a clinical trial is often complex, as it was pointed out by the organizers of the shared task. For example, one of the criteria for "Advanced-CAD" is "Taking two or more medications to treat CAD". It involves three steps – knowing CAD medication (brand and generic) names, recognizing the patient is presently taking them, and the reason for the medications was to treat CAD not simply for prevention. There were also other complexities such as in "patient speaks English", the language status is rarely discussed in clinical notes and therefore must be inferred from mention of the presence of an interpreter.

The NLP system described here was developed to address the complexities of the shared task and to participate in the challenge. Our system was statistically tied for the first place (along with 7 others) out of 45 participants. Subsequent to the challenge, noticing the importance of the knowledge-resources accuracy, we developed a novel, automated process of curating knowledge resources. It enabled the system to improve performance (even though the headroom was small to begin with) and achieved numerically the highest performance. In addition, the automated process was greatly beneficial in reducing complexity and time to develop the methods (E.g. "Alcohol Abuse" originally consisted of painstaking hand-crafting of lexicon; with this system we were able to rapidly generate a lexicon and rely solely on it for good results). Even more significantly, the resource curation approach is useful beyond this cohort selection task, especially to phenotype extraction and classification tasks.

## Methods

We approached each of the thirteen cohort-selection criteria individually, with either a rule-based or a supervised learning approach, and often used word-embeddings (unsupervised training of word representations) to enhance lexicons. For all criteria, we started with a rule-based approach but for some either the complexity of the rules or performance was the impetus to convert the rules into features and use a supervised learning model. In some cases, after building both methods, we chose the rules due to better performance. The supervised learning methods were trained using the gold standard provided.

## Cohort Selection Criteria

As shown in Table 1, the thirteen cohort-selection criteria are wide ranging. They involve lab test results, present medical problems, past medical history, current treatments and the reason for the treatments, use of over the counter medications, drugs/alcohol consumption, and language spoken.

*Table 1. Thirteen cohort selection criteria that were defined in the N2C2 Task 1 challenge.*

| Num. | Short Names | Description |
| --- | --- | --- |
| 1 | Drug Abuse | Drug abuse, past or present |
| 2 | Alcohol Abuse | Current alcohol use over weekly recommended limits |
| 3 | English | Patient (predominantly) speaks English |
| 4 | Makes-Decisions | Patient makes their own medical decisions |
| 5 | Abdominal | History of intra-abdominal surgery, small or large intestine resection or small bowel obstruction |
| 6 | Major-Diabetes | Any of the following that are a result of (or strongly correlated with) uncontrolled diabetes: Amputation, Kidney damage, Skin conditions, Retinopathy, Nephropathy, or Neuropathy |
| 7 | Advanced-CAD | Advanced cardiovascular disease having two or more of:<br>• Taking two or more medications to treat CAD<br>• History of myocardial infarction<br>• Currently experiencing angina<br>• Ischemia, past or present |
| 8 | MI-6Months | Myocardial infarction in the past 6 months |
| 9 | Keto-1Yr | Diagnosis of ketoacidosis in the past year |
| 10 | DietSupp-2Months | Supplements (excluding Vitamin D) in the past 2 months |
| 11 | ASP-for-MI | Use of aspirin to prevent myocardial infarction |
| 12 | HBA1c (high) | Any HbA1c value between 6.5 and 9.5% |
| 13 | (Elevated) Creatinine | Serum Creatinine > normal upper limit |

## Dataset

The dataset consisted of 3 to 5 for clinical notes for each patient, which are ordered in time but they are not necessarily consecutive. Clinical notes had date and time stamps which were systematically shifted into future by a (patient-specific) random amount of time. The date of the latest clinical note was to be considered as the *present day* for temporal criteria or sub-criteria. The training set consisted of clinical notes for 202 patients. Each patient record was annotated with patient level tags indicating if the patient "met" or "not met" each of the 13 criteria. Clinical notes of ten of the patients were further annotated with (non-exhaustive) detailed evidence of the patient-level annotation. For example, if a patient in this smaller set were to have MI in the past 6 months, a specific sentence indicating the fact was annotated as detailed evidence. The test set consisted of 3-5 clinical notes for additional unannotated 86 patients, which was released a couple of days before the deadline for results submission.

### Basic biomedical NLP processing

As a first step we processed each clinical note in the dataset using the CLAMP biomedical NLP software [5] with the built-in "NER-attribute" pipeline. In addition to tokenization and sentence segmentation, it provided medical named entity recognition, entity mapping to standard medical coding systems (e.g. SNOMED and RxNORM), section identification, and assertions such as presence and negation. Most but not all of our analytics leveraged the output of this preprocessing step.

### General framework

For each criterion and sub-criterion there were three steps in developing the method for it:
1. Determine knowledge resources needed and build them using available information from the Web or via sampling the pre-trained biomedical word2vec model as criterion-specific lexicons (e.g. names of the drugs used to treat CAD);
2. Develop a rule-based method to determine if the patient meets the criterion or sub-criterion or not (e.g. Identify occurrences of the CAD drugs in a patient's clinical notes)
3. Ensure additional constraints are considered (e.g. the patient is indeed taking the drugs, i.e. not negated nor hypothetical, and that the drugs prescribed for CAD for the patient).

For each such method, there were multiple iterations of fine-tuning rules (using the training data) and when the rules became too complex, the rules were replaced with a supervised learning model, which were tuned via cross-validation performed on the training data. Even more importantly, there was a significant effort to improve the accuracy of the knowledge resources, which was initially done through trial and error and was later refined into a systematic, automated method that will be described below. The knowledge resources were lists of words or phrases, and therefore in the linguistic sense they were task-specific lexicons. The draft versions of the lexicons were obtained from the Web or manually created, and then the word2vec model pre-trained on biomedical text [6] was used for expanding it with similar words and phrases. The overall system schematic is shown in Figure 1.

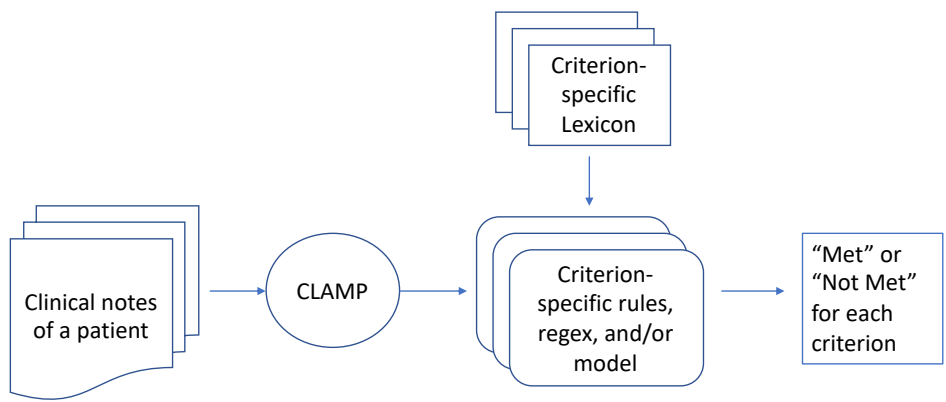

*Figure 1. A schematic of our overall system for cohort selection.*

## Methods for Each Criterion

<u>Drug-Abuse</u>: BioNLP preprocessing (CLAMP) output was used for this criterion in order to extract entities from the text that were "problem" or "drug" entities, and to determine if these entities were mentioned in the patient history section. A small list of normally abused drugs (lexicon) was manually constructed from searching the Web, and the drug mentions were matched with the list. Rules checked to ensure the drugs were not prescribed using a heuristic that the presence of dosage of the drug implies prescribed use. CLAMP's assertion identifier was also used to rule out negated mentions of phrases indicating drug abuse, which accurately asserted whether the criterion was met or not met. The complexity with this criterion was threading the needle between identifying whether the mentioned drugs were abused or prescribed; this was accomplished through a combination of drug dosage detection and assertion detection (using CLAMP) along with regex rules.

<u>Alcohol-Abuse:</u> Various expressions, relating to alcohol abuse, binge drinking, or alcohol abuse-related terms such as 'AA' (Alcoholic Anonymous), were constructed as rules using manually curated lexicon; their presence indicated "met". One interesting component of this criteria was determining what would constitute "abuse" — in particular, mention of another family member expressing concern about the patient's drinking was a good indicator of abuse, and thus was used in rules. Additional temporal constraints were used in the rules to disqualify mentions of abuse that clearly occurred in the past or that had been stopped. CLAMP output was not needed in the method for this criterion.

<u>English speaking:</u> A list of languages was obtained from an online source (http://aboutworldlanguages.com/languages-a-z), and text search was conducted for occurrences of language mentions other than English, in combination with words/phrases indicating spoken language; the heuristic was that a match indicated this criterion was not met. Additionally, a text search for the mention of an interpreter was conducted; if found (considering possible negation), the criteria was not met. We trained a machine learning model with the aforementioned factors as features, but there was no performance improvement. This criterion was presumed to be "met" by default, and only changed to "not met" in the presence of the information mentioned above.

<u>Makes Decisions</u>: A Naive Bayes classifier was trained and cross-validated on the training set. Features were the number of times each of the following regex (in python syntax) occurred in the raw text:

*/dementia/*
*/(?<!psychomotor )retard/*
*/(altered mental|mental stat)/*

The first regex obviously matches the sequence *dementia*; the second matches occurrences of *retard* that does not follow *psychomotor*; and the last regex matches *altered mental* or *mental stat* patterns. We assumed not-met unless the model returns a score above a high threshold (determined heuristically).

<u>Abdominal:</u> CLAMP output was used to obtain medical problems mentioned in the "Past Medical History" and "History of Present Illness" sections. These extracted problems were matched with a dictionary consisting of terms indicating abdominal surgeries and bowel obstruction. The dictionary (the lexicon) was compiled manually by curating names of common surgical procedures from Wikipedia. We learned (from the training data) that some procedures, such as 'aortic valve replacement', although not generally thought of as abdomen related, involve surgery that falls under the criterion, and hence was added to the lexicon.

Major-Diabetes: For this criterion, we used medical concepts which were tagged as problem and treatment by the BioNLP preprocessing software. A support vector machine-based classification model was trained. Count of terms in the patient record that indicate the following clinical characteristics of patients which are not in the family history section were used as features in the model:
1. bad skin conditions
2. kidney damage
3. neuropathy
4. nephropathy
5. retinopathy

Terms (lexicons) related to bad skin conditions, kidney damage, neuropathy, nephropathy, and retinopathy were curated using Google and pre-trained biomedical embeddings.

Advanced-CAD: We used CLAMP output for drugs, problems, and assertions on them in the patient record, and to exclude entities in the social history. Rules were used to test for:
- Presence of ischemia (assertion states "present");
- MI-related words (assertion states "present");
- Angina and angina-related terms in the most recent clinical note and assertion states "present"; and
- Presence of (two or more) medications for CAD and that the patient has CAD.

The CAD-Medication list was manually curated from online angina medications list [7] from Healthline, and from general CAD information and medications [8], beta-blockers [9], and ACE-inhibitors [10] from the online WebMD pages. An important consideration was that even if angina was mentioned to be present in the recent clinical note, if it was later negated then the final decision should be "not met" for the "currently experiencing angina" sub-criterion.

MI-6Months: The 'section' tags and medical problems from the CLAMP output were used. Synonymous MI terms such as "NSTEMI", "STEMI", "NQWMI", "IMI", "myocardial", and "myocardial infarction" were manually curated and used to determine if patient had MI. Negated mentions and mentions in the past medical history section were discarded. The timestamps on the clinical notes were once source of the temporal assertion on the MI mentions. Another source was the mentions of temporal events such as "9 months ago" in the sentence containing the MI term. HeidelTime [11] was used for interpreting time labels in a consistent way, however, further specialization was needed with partial date specification such as "3/78".

Keto-1Yr: This was not present in the entire training set and so we created simple rules to test for the presence of "ketoacidosis" (its variants) as a disease and/or positive urine-ketones test

(and its variants). Later, we learned that there was not a positive example in the test dataset either.

DietSupp-2Months: A resource of generic names and brand names of dietary supplements was developed using resources from MedlinePlus [12]. The named entities of types treatments and medications from the CLAMP output were checked against it using two-way substring matching after conversion to all lower case. The two-way substring matching returns true if s1 is a substring of s2 or vice-versa.

ASP-for-MI: For this category the CLAMP output was used to identify terms which the software identified as a problem, a treatment, or a drug. A Decision Tree based classification model performed better than a logistic regression or SVM for this task. The following features were used in the classification model:
1. If the patient was prescribed Aspirin or not: This was determined by searching for aspirin and other medical names of aspirin (such as ASP) in the CLAMP output for drugs and treatments.
2. Dosage of the prescribed Aspirin (or its variants).
3. If the patient was diagnosed with MI or not: This feature was also determined by searching for not only MI but also for medical synonyms of MI in the CLAMP output.
4. Whether the patient had symptoms indicating MI: The terms indicating the symptoms were compiled using an initial lexicon manually curated from online sources and later expanded as below.

The lexicons for Aspirin, MI, and MI symptoms were expanded using pre-trained biomedical word2vec model [6] and manual inspection.

HBA1C (high): The biomedical NLP pre-processing software, CLAMP, was used to extract keywords which were tagged as either lab tests or lab values. A support vector machine classification model performed best in determining whether the input document satisfies the HBA1c criteria or not. The following features were extracted from the CLAMP output using regular expressions for this model:
1. Presence of keywords related to hemoglobin in the document: The keywords similar to hemoglobin were retrieved from the pre-trained Google embeddings from newswire articles [13] and (separately) from the embeddings from biomedical text [6].
2. Presence of a lab value corresponding to keywords related to hemoglobin.
3. Range of the lab values associated with hemoglobin or similar terms: Range was extracted from the text tagged as lab value using regular expressions.

(Elevated) Creatinine: Regular expressions and rules were used with the following arguments:
1. *Elevated Creatinine* or words similar to it were present in the clinical notes;
2. Lab test values (either listed as a series of lab name and value pairs, or in a table format, or mentioned in the text);
3. Applying the range given in the clinical note to creatinine values in the note, otherwise applying the generally accepted normal ranges (see below); and
4. Ignoring creatinine levels taken from urine analysis.

According to Wikipedia, generally acceptable normal levels of creatinine in the blood are 0.6 to 1.2 milligrams per deciliter (DL) in adult males and 0.5 to 1.1 milligrams per deciliter in adult females. In our rules, the criterion was "not met" if the values were in the normal range specified in the clinical note or within general normal range plus 0.5 (to account for the variability in what is generally considered "normal"); otherwise the criterion was "met". Neither external resources nor CLAMP output were used.

## A Novel Approach for Lexicon Curation

As can be seen from the description of the methods above, there was a frequent need for task-specific lexicon, a knowledge resource. Although we used manual curation for the Challenge (due to the time constraints), it became obvious to us that there was a critical need for a standardized and automated process. Therefore, after the Challenge, we turned our attention to developing such method to obtain task specific lexicons, for use in rule- or classifier-based solutions in general. The resulting novel approach consists of the following steps:

1. Create n-grams (up to n=4) from the training data; use these as features in a logistic regression model;
2. Train and test the logistic regression model using 5-fold cross-validation on the training set; store average coefficient weights corresponding to each "feature" (n-gram) from the models;
3. From the model (averaged over cross-validation steps), identify n-grams corresponding to the most positive and negative coefficients; We hypothesized that these are the most discriminative features. Keep the n-grams whose absolute coefficients are above a threshold (which is also learned through cross-validation);
4. Augment this vocabulary with terms within a small similarity distance (determined by trial and error) to n-grams obtained in step 3 above in a pre-trained word2vec model on biomedical text [6]; iterate upon terms for a few times (also determined heuristically) to expand the list further;
5. Use the combination of logistic regression-obtained and word2vec-obtained terms as the lexicon in rules- or classifier-based system; and
6. If wanting to expand/narrow vocabulary, carry out steps 1-5 with differing thresholds for the coefficient values in step 3 and the similarity distance in the word embeddings in step 4.

This system represents lexicon selection from data within the training set (internally cross-validated on the training dataset) and from related data outside the training set to catch any similar terms that may not be in the training set but are still useful (i.e. externally enhanced using word embedding pre-trained on related corpora). We refer to this system as the Internal plus External Lexicon Selection (IELS) method. It should be noted that this approach not only selects single words (unigrams) but also multi-word phrases (n-grams) which are often common in biomedical text. This approach is a variant of well-known dimensionality reduction methods such as PCA (Principal Component Analysis), PMI (Point-wise mutual information) based feature selection, and LASSO (L1 regularized least squares regression). We applied this approach in rebuilding lexicons for a few of the criteria.

## Performance metrics

Accuracy of the system was measured for each criterion as well as together for all criteria. Standard precision (P), recall (R), F measure, and Area under the precision-recall curve (AUC) were used. Every decision made by the system for each criterion was scored as one of True Positive, True Negative, False Positive, or False Negative (i.e. no multiple counting of errors). For the aggregate measure, both micro and macro average were calculated. The scripts provided by the Challenge organizers were used in reporting the metrics.

*Table 2. Performance accuracy of our system as submitted to the N2C2 Challenge*

```
******************************** TRACK 1 ********************************
              ------------ met --------------    ------ not met -------   -- overall ---
              Prec.   Rec.    Speci.  F(b=1)     Prec.   Rec.    F(b=1)    F(b=1)  AUC
    Abdominal 0.9231  0.8000  0.9643  0.8571     0.9000  0.9643  0.9310    0.8941  0.8821
  Advanced-cad 0.8182  0.8000  0.8049  0.8090    0.7857  0.8049  0.7952    0.8021  0.8024
  Alcohol-abuse 0.0000 0.0000  0.9880  0.0000    0.9647  0.9880  0.9762    0.4881  0.4940
    Asp-for-mi 0.8400  0.9265  0.3333  0.8811    0.5455  0.3333  0.4138    0.6475  0.6299
     Creatinine 0.7407  0.8333  0.8871  0.7843   0.9322  0.8871  0.9091    0.8467  0.8602
   Dietsupp-2mos 0.8511 0.9091 0.8333  0.8791    0.8974  0.8333  0.8642    0.8717  0.8712
      Drug-abuse 0.6667 0.6667 0.9880  0.6667    0.9880  0.9880  0.9880    0.8273  0.8273
         English 0.9605 1.0000 0.7692  0.9799    1.0000  0.7692  0.8696    0.9247  0.8846
           Hba1c 0.9667 0.8286 0.9804  0.8923    0.8929  0.9804  0.9346    0.9134  0.9045
         Keto-1yr 0.0000 0.0000 1.0000 0.0000    1.0000  1.0000  1.0000    0.5000  0.5000
   Major-diabetes 0.9394 0.7209 0.9535 0.8158    0.7736  0.9535  0.8542    0.8350  0.8372
   Makes-decisions 0.9880 0.9880 0.6667 0.9880   0.6667  0.6667  0.6667    0.8273  0.8273
          Mi-6mos 0.3750 0.7500 0.8718 0.5000    0.9714  0.8718  0.9189    0.7095  0.8109
              ----------------------------       ---------------------     ---------------
Overall (micro) 0.8807  0.8845  0.9165  0.8826   0.9193  0.9165  0.9179    0.9003  0.9005
Overall (macro) 0.6976  0.7095  0.8493  0.6964   0.8706  0.8493  0.8555    0.7759  0.7794
```

## Results

Table 2 shows accuracy metrics for the system that was submitted to the Challenge. The overall micro accuracy was 0.9003 and AUC was 0.9005. The test data is highly skewed, for example, Alcohol abuse, Drug abuse, Keti-1Yr, and MI-6mos have less than 7 positive cases, and Makes-Decisions has only 3 negative instances. Therefore, the F measure for the individual criteria varied in a wide range from 0.4881 to 0.9247. Excluding these skewed criteria, F measure was in a tighter range of 0.8303 to 0.9247 (except for ASP-for-MI, which was 0.6475), indicating a good balance of performance among our methods.

Table 3 shows changes in the micro-averaged F measure with the systematic lexicon curation for Major-Diabetes, Drug-abuse, and Alcohol-abuse. The goal of this approach is a systematic methodology to otherwise laborious and error-prone lexicon curation and therefore small or no improvement in accuracy is acceptable. Further studies are needed to test its effectiveness.

*Table 3. Changes in accuracy with the model-based lexicon curation*

| Criterion | F Measure | |
|---|---|---|
| | With manual curation | With the model-based curation |
| Major-Diabetes | 08350 | 0.8350 |
| Drug-abuse | 0.8273 | 0.8273 |
| Alcohol-abuse | 0.4881 | 0.4911 |

Table 4 shows improved accuracy metrics for the system after the Challenge. The improvements were achieved using the systematic lexicon creation for the three criteria as well as debugging of the code and regular expressions (only using the training data). Improvement was seen in Advanced-CAD, Alcohol-abuse, Creatinine, and MI-6months. These changes resulted in the improvement of the overall micro-averaged F measure to 0.9140 and the overall AUC to 0.9139. Numerically, the overall F measure is better than the best score at the Challenge. The top overall F measures along with our new accuracy score are shown in Table 5.

*Table 4. Performance accuracy of our system with improvements after the Challenge.*

```
*************************************** TRACK 1 ***************************************
              ------------- met -------------    ------- not met --------   -- overall ---
              Prec.   Rec.    Speci.  F(b=1)     Prec.   Rec.    F(b=1)     F(b=1)  AUC
   Abdominal  0.9231  0.8000  0.9643  0.8571     0.9000  0.9643  0.9310     0.8941  0.8821
 Advanced-cad 0.8039  0.9111  0.7561  0.8542     0.8857  0.7561  0.8158     0.8350  0.8336
 Alcohol-abuse 0.0000 0.0000  1.0000  0.0000     0.9651  1.0000  0.9822     0.4911  0.5000
   Asp-for-mi 0.8400  0.9265  0.3333  0.8811     0.5455  0.3333  0.4138     0.6475  0.6299
   Creatinine 0.9091  0.8333  0.9677  0.8696     0.9375  0.9677  0.9524     0.9110  0.9005
 Dietsupp-2mos 0.8511 0.9091  0.8333  0.8791     0.8974  0.8333  0.8642     0.8717  0.8712
   Drug-abuse 0.6667  0.6667  0.9880  0.6667     0.9880  0.9880  0.9880     0.8273  0.8273
      English 0.9605  1.0000  0.7692  0.9799     1.0000  0.7692  0.8696     0.9247  0.8846
        Hba1c 0.9667  0.8286  0.9804  0.8923     0.8929  0.9804  0.9346     0.9134  0.9045
      Keto-1yr 0.0000 0.0000  1.0000  0.0000     1.0000  1.0000  1.0000     0.5000  0.5000
 Major-diabetes 0.9394 0.7209 0.9535  0.8158     0.7736  0.9535  0.8542     0.8350  0.8372
 Makes-decisions 0.9880 0.9880 0.6667 0.9880     0.6667  0.6667  0.6667     0.8273  0.8273
      Mi-6mos 0.5833  0.8750  0.9359  0.7000     0.9865  0.9359  0.9605     0.8303  0.9054
              ----------------------------       ------------------------   ---------------
Overall (micro) 0.8996 0.8976 0.9302  0.8986     0.9288  0.9302  0.9295     0.9140  0.9139
Overall (macro) 0.7255 0.7276 0.8576  0.7218     0.8799  0.8576  0.8641     0.7929  0.7926
```

## Discussion

Cohort selection is similar to phenotype extraction and classification, since as described in a review [14] of approaches to identifying phenotype cohorts using electronic health records, phenotype may be broadly defined as the observable characteristic of a patient. The review concluded that the majority of the studies used rules and regular expressions, for simplicity as well as for their effectiveness. We noticed the same in developing our own system. In previous studies, typically the rules were developed based on clinician description and judgement as in the system [15] that analyzed discrete data and HL7 text messages for identifying eligible patients for cancer trials. Some studies developed rules from guidelines (for example, to screen patients at risk for type 2 diabetes [16]) or refined existing rules. In our case, we used the annotation guidelines provided by the Challenge and our own understanding of the criteria, based on the online and text book knowledge, to build and refine the rules.

Several off-the-shelf biomedical NLP systems such as cTAKES [17], MedLEE [18], and Metamap [19] are available to identify key medical concepts from clinical text. An earlier study [20] even went one step further and tried to extract relevant sentences from clinical notes given clinical trial eligibility (using the same dataset that was used here from an earlier challenge), with the expectation that these sentences help narrow the scope of the rules. But the accuracy was low. We found that CLAMP [5] met our requirements well in terms of providing the necessary concepts, assertions, and section tags, and thus allowed us to focus on criterion-specific logic and lexicons.

*Table 5. Results for the top ten teams at the challenge (shaded entries statistically tied for the 1st place) and our post-submission accuracy.*

| Team name | Overall micro F measure |
|---|---|
| Arizona State University (post-submission) | 0.9140 |
| MedUniGraz | 0.9100 |
| University of Michigan | 0.9075 |
| Sorbonne Universite | 0.9069 |
| Med Data Quest | 0.9028 |
| Cincinnati Children's Hospital Medical Center | 0.9026 |
| Arizona State University (original submission) | 0.9003 |
| University of NSW / National Cancer Institute | 0.8913 |
| Harbin Institute of Technology | 0.8855 |
| University of Utah | 0.8837 |
| NTTMUNSW | 0.8765 |

The TREC medical record track conducted challenges for cohort selection in 2011 and 2012 as *information retrieval* (IR) tasks [21]. Because the goal was to find the most relevant documents (e.g. clinical notes), the task and performance of the participating systems are not comparable to our system or this task. However, an analysis of the barriers in retrieving relevant information from patient records [22] and the use of the physician judgements to learn relevance models [23] are in general relevant to cohort selection from patient records.

In the present cohort selection challenge, a total of 45 teams participated, and most of the top-performing systems predominantly used rules, regular expressions, and off-the-shelf concept extraction biomedical software (https://portal.dbmi.hms.harvard.edu/projects/n2c2-t1/). However, the systems did make use of simple learning systems for narrow purposes. In fact, systems that exclusively relied on learning methods performed poorly, indicating the need for fine tuning the analytics in this task and the limited size of the training and test sets. Our general approach is therefore similar to the other top performing systems, however, we emphasize the use and development of lexicon because of its importance in cohort selection.

## Conclusion

This paper reported a system we developed for the N2C2 cohort selection shared challenge and our system was statistically tied for the first place along with 7 other systems among a total of 45 participants. We took a divide and conquer approach by developing separate analytics for each of the 13 criteria for cohort selection. The methods usually but not always used the output of a biomedical preprocessing software, CLAMP, such as medical concepts, asserts, and section tags. Most analytics used rules and regular expressions but three analytics used supervised learning (SVM and decision trees). All of them critically depended on knowledge resources in the form of lexicons relevant to the criteria. Therefore, subsequent to the challenge we developed a supervised learning model-based approach to curating lexicons. The key benefit of this approach is to automate and standardize the curation process. Additional tuning of the rules resulted in performance that is numerically higher than the best performance from the Challenge. A challenge for the future is to develop automatic methods that integrate domain knowledge (that are now used in manual creation of rules) with machine learning methods.

## Acknowledgement
We thank Dr. S. Soumya Panchanathan, MD MS, for helping us understand clinical nuances around diagnosis and treatment of some of the medical conditions in the cohort selection.
## Conflict of Interest
The authors do not have any conflicts of interest.